\documentclass[letterpaper, 10 pt, conference]{ieeeconf}  

\IEEEoverridecommandlockouts                              

\usepackage{cite}
\usepackage{amsmath,amssymb,amsfonts}
\usepackage{algorithm,algpseudocode}
\usepackage{graphicx}
\usepackage{textcomp}
\usepackage{xcolor}
\usepackage{multirow}
\usepackage{subcaption}
\usepackage{float}
\usepackage{multirow}
\usepackage{enumerate}
\usepackage{subcaption}
\usepackage{textcomp}
\usepackage[algo2e]{algorithm2e}
\DeclareGraphicsExtensions{.pdf,.png,.jpg,.tif}
\usepackage{balance}

\title{\LARGE \bf
An Enhanced Differential Evolution Algorithm Using a Novel Clustering-based Mutation Operator
}

\author{Seyed Jalaleddin Mousavirad$^{1}$, Gerald Schaefer$^{2}$, Iakov Korovin$^{3}$, Mahshid Helali Moghadam$^{4,5}$,\\ Mehrdad Saadatmand$^{5}$ and Mahdi Pedram$^{6}$
\thanks{This work was supported by RFBR  grant 20-04-60485.}%
\thanks{$^{1}$S. J. Mousavirad is with the Computer Engineering Department, Hakim Sabzevari University, Sabzevar, Iran}%
\thanks{$^{2}$G. Schaefer is with the Department of Computer Science, Loughborough University, Loughborough, U.K.}%
\thanks{$^{3}$I. Korovin is with Southern Federal University, Taganrog, Russia}%
\thanks{$^{4}$M. H. Moghadam is with M\"alardalen University, V\"aster\'as, Sweden}%
\thanks{$^{5}$M. H. Moghadam and M. Saadatmand are with RISE Research Institutes of Sweden, Sweden}%
\thanks{$^{6}$M. Pedram is with Lorestan University of Medical Sciences, Khorramabad, Iran}%
}
\begin{document}

\maketitle
\thispagestyle{empty}
\pagestyle{empty}

\begin{abstract}
Differential evolution (DE) is an effective population-based metaheuristic algorithm for solving complex optimisation problems. However, the performance of DE is sensitive to the mutation operator. In this paper, we propose a novel DE algorithm, Clu-DE, that improves the efficacy of DE using a novel clustering-based mutation operator. First, we find, using a clustering algorithm, a winner cluster in search space and select the best candidate solution in this cluster as the base vector in the mutation operator. Then, an updating scheme is introduced to include new candidate solutions in the current population. Experimental results on CEC-2017 benchmark functions with dimensionalities of 30, 50 and 100 confirm that Clu-DE yields improved performance compared to DE.
\end{abstract}

\section{Introduction}
Optimisation problems exist in a variety of scientific fields, varying from medicine to agriculture. While conventional optimisation algorithms are popular, they suffer from drawbacks such as getting stuck in local optima and being sensitive in the initial state~\cite{GHMS-RCS}. To tackle these problems, population-based metaheuristic algorithms such as particle swarm optimisation~\cite{PSO_Main_Paper02} offer a powerful alternative thanks to their well-recognised characteristics, such as self-adaptation and being derivative free~\cite{center_sampling_PSO_SMC2020}. 

Differential evolution (DE)~\cite{DE_Original} is a simple yet effective population-based algorithm which has shown good performance in solving optimisation problems in areas including image processing~\cite{Image_Segmentation_auto_DE,Image_Thresholding_CenDE}, pattern recognition~\cite{ANN_DE_opposition,ANN_RDE-OP,ANN_DE_cen_Op}, and economics~\cite{DE_financial01,DE_financial02}. DE is based on three primary operators: mutation, which generates new candidate solutions based on scaling differences among candidate solutions, crossover, which combines a mutant vector with the parent one, and selection, which selects a better candidate solution from a new one and its parent.

The performance of DE is directly related to these operators~\cite{CI-DE}. Among them, the mutation operator plays a crucial role to generate new promising candidate solutions and significant recent work has focussed on developing effective mutation operators. \cite{DE_ensemble02} proposes a multi-population DE which combines three different mutation strategies including current-to-pbest/1, current-to-rand/1, and rand/1. \cite{CODE_01} employs three trial vector generation strategies and three control parameter settings and randomly selects between them to create new vectors. In~\cite{DE_tournament}, $k$-tournament selection is used to introduce selection pressure for selecting the base vector. \cite{DE_neighborhood} proposes a neighbourhood-based mutation that is performed within each Euclidean neighbourhood. In~\cite{DE_Competition_ICCSE2019}, a competition scheme for generating new candidate solutions is introduced so that candidate solutions are divided into two groups, losers and winners. Winners create new candidate solutions based on standard mutation and crossover operators, while losers try to learn from winners.

In this paper, we propose a novel DE algorithm, Clu-DE, which employs a novel clustering-based mutation operator. Inspired by the clustering operator in the human mental search (HMS) optimisation algorithm~\cite{HMS_Main_Paper}, Clu-DE clusters the current population into groups and selects a promising region as the cluster with the best mean objective function value. The best candidate solution in the promising region is selected as the base vector in the mutation operator. An updating strategy is then employed to include the new candidate solutions into the current population.  Experimental results on CEC-2017 benchmark functions with dimensionalities of 30, 50 and 100 confirm that Clu-DE yields improved performance compared to DE.

The remainder of the paper is organised as follows. Section~\ref{Sec:DE} describes the standard DE algorithm and some preliminaries about clustering. Section~\ref{Sec:proposed} introduces our Clu-DE algorithm, while Section~\ref{Sec:Exp} provides experimental results. Section~\ref{Sec:conc} concludes the paper.

\section{Background}
\subsection{Differential Evolution}
\label{Sec:DE}
Differential evolution (DE)~\cite{DE_Original} is a simple but effective population-based optimisation algorithm based on three main operators: mutation, crossover, and selection.

The mutation operator generates a mutant vector $\overrightarrow{v_{i}}=(v_{i,1},v_{i,2},...,v_{i,D})$ for each candidate solution as
\begin{equation}
\overrightarrow{v_{i}}=\overrightarrow{x_{r1}}+F(\overrightarrow{x_{r2}}-\overrightarrow{x_{r3}}) ,
\end{equation} 
where $\overrightarrow{x_{r1}}$, $\overrightarrow{x_{r2}}$, and $\overrightarrow{x_{r3}}$ are three distinct candidate solutions randomly selected from the current population and $F$ is a scaling factor.

Crossover shuffles the mutant vector with the parent vector. For this, binomial crossover, defined as
\begin{equation}
u_{i,j}=\begin{cases}
v_{i,j} & \mbox{if }rand(0,1)\leq CR  \mbox{ or } j==j_{rand} \\
x_{i,j} & \text{otherwise}
\end{cases} 
\end{equation}
is employed, where $\overrightarrow{u_{i}}$ is called a trial vector, $CR$ is the crossover rate, and $j_{rand}$ is a random integer number between 1 and the number of dimensions. 

Finally, the selection operator selects the better candidate solution from the new candidate solution and its parent to be passed to the new population.

\subsection{Clustering}
Clustering is an unsupervised pattern recognition technique to partition samples into different groups so that the members of a cluster share more resemblance compared to members of different clusters. $k$-means~\cite{k-means_Original} is the most popular clustering algorithm based on a similarity measure (typically Euclidean distance). It requires to define $k$, the number of clusters, in advance and proceeds as outlined in Algorithm~\ref{Alg1}. 

\begin{algorithm2e}[h!]
	\SetAlgoLined
	\SetKwInOut{Input}{Input}\SetKwInOut{Output}{Output}
	\Input{ $X$: samples\\ \ $k$: number of clusters}
	\Output{ $C={c_{1},c_{2},...,c_{k}}$: the set of clusters defined by their centres $\mu_j$}
	\BlankLine
	Initialise cluster centres $\mu_{1}, \mu_{2},...,\mu_{k}$ randomly \;
	\While{stopping condition is not met}{
		Assign each sample $x_i$ to the closest cluster centre, i.e.\ based on $\underset{j}{argmin} ||x_{i}-\mu_{j}||$ \;
		For all clusters, calculate new cluster centre as $\mu_{j}=\frac{(\sum_{c_{i}=j} x_{i})}{n_{j}}$, where $n_{j}$ is the number of samples in the $j$-th cluster \;
	}
	\caption{Pseudo-code of $k$-means clustering algorithm.}
	\label{Alg1}
\end{algorithm2e}

\section{Proposed Clu-DE Algorithm}
\label{Sec:proposed}
In this paper, we improve DE using a novel clustering-based mutation and updating scheme. Our proposed algorithm, Clu-DE, is given, in the form of pseudo-code in Algorithm~\ref{Alg2}, while in the following we describe its main contributions.

\begin{algorithm2e}[t!]
	\SetAlgoLined
	\SetKwInOut{Input}{Input}\SetKwInOut{Output}{Output}
	\Input{ $D$: dimensionality of problem\\ \ $NFE_{\max}$: max.\ no.\ of function evaluations\\ \ $N_{P}$: population size\\ \ $F$: scaling factor\\ \ $C_{R}$: crossover probability\\ \ $M$: no.\ of new candidate solutions}
	\Output{ $x^*$: the best solution }
	\BlankLine
	Generate initial population $Pop$ uniform randomly\;
	Calculate objective function value (OFV) for each candidate solution \;
	$NFE=N_{P}$ \;
	\While{$NFE<NFE_{\max}$}{
		\For {$i\leftarrow 1$ \KwTo $N_{P}$}{ 		
			Select three parents, $x_{i1}$, $x_{i2}$, and $x_{i3}$, randomly from current population with $x_{i1} \neq x_{i2} \neq x_{i3}$ \; 
			$ v_{i}=x_{1}+F(x_{i2}-x_{i3})$ \;
			\For {$j\leftarrow 0$ \KwTo $D$}{
				\eIf{$rand_{j}[0,1]<C_{R}$ or $j==j_{rand}$} {$u_{i,j}={v}_{i,j}$}
				{$u_{i,j}=x_{i,j}$}
			} 
			Calculate OFV of $u_{i}$\;
			\eIf{$f(u_{i})<f(x_{i})$}
			{$ \bar{x} \leftarrow u_{i} $ \;}
			{$ \bar{x} \leftarrow x_{i} $ \;} 
			$Pop(i) \leftarrow \bar{x} $\;
		}
		$k$ = random number between 2 and $\sqrt{N_{P}}$ \;
		Cluster $N_{P}$ candidate solutions into $k$ clusters\;
		Calculate mean OFV of each cluster\;
		winner cluster = cluster with lowest mean OFV \;
		$winner$ = best bid in winner cluster \;
		\For {$j\leftarrow 1$ \KwTo $M$}{
			Select two parents, $x_{i1}$ and $x_{i2}$, randomly from current population with $x_{i1} \neq x_{i2}$ \; 
			$v^{clu}_{j}=winner+F*(x_{i1}-x_{i2})$ \;
		}
		Select $M$ candidate solutions randomly from current population as set $B$\;
		From $v^{clu} \cup B$, select  best $M$ individuals as $\bar{B}$\;
		Update new population as $(P-B) \cup \bar{B}$. \;
		$NFE=NFE+N_{P}$ \;	
	}
	$x^{*} \leftarrow$ the best candidate solution in $Pop$ 
	\caption{Pseudo-code of Clu-DE algorithm.}
	\label{Alg2}
\end{algorithm2e}

\begin{table}[b!]
\centering
\setlength{\tabcolsep}{3pt}
\caption{Summary of CEC2017 benchmark functions~\cite{CEC2017}. $N$ indicates the number of basic functions to form hybrid and composite functions. The search range is $[+100,-100]^{D}$ in all cases.}
\label{tab:func}
\begin{tabular}{lll}
\hline
\multicolumn{3}{l}{Unimodal functions} \\
&F1 & Shifted and Rotated Bent Cigar Function \\
&F2 & Shifted and Rotated Sum of Different Power Function \\
&F3 & Shifted and Rotated Zakharov Function \\
\hline
\multicolumn{3}{l}{Multimodal functions} \\
&F4 & Shifted and Rotated Rosenbrock's Function \\
&F5 & Shifted and Rotated Rastrigin's Function \\
&F6 & Shifted and Rotated Expanded Scaffer's Function \\
&F7 & Shifted and Rotated Lunacek Bi\_Rastrigin Function \\
&F8 & Shifted and Rotated Non-Continuous Rastrigin's Function \\
&F9 & Shifted and Rotated Levy Function \\
&F10 & Shifted and Rotated Schwefel's Function \\
\hline
\multicolumn{3}{l}{Hybrid multimodal functions} \\
&F11 & Hybrid Function 1 ($N=3$) \\
&F12 & Hybrid Function 2 ($N=3$) \\
&F13 & Hybrid Function 3 ($N=3$) \\
&F14 & Hybrid Function 4 ($N=4$) \\
&F15 & Hybrid Function 5 ($N=4$) \\
&F16 & Hybrid Function 6 ($N=4$) \\
&F17 & Hybrid Function 7 ($N=5$) \\
&F18 & Hybrid Function 8 ($N=5$) \\
&F19 & Hybrid Function 9 ($N=5$) \\
&F20 & Hybrid Function 10 ($N=6$) \\
\hline
\multicolumn{3}{l}{Composite functions} \\
&F21 & Composition   Function 1 ($N=3$) \\
&F22 & Composition   Function 2 ($N=3$) \\
&F23 & Composition   Function 3 ($N=4$) \\
&F24 & Composition   Function 4 ($N=4$) \\
&F25 & Composition   Function 5 ($N=5$) \\
&F26 & Composition   Function 6 ($N=5$) \\
&F27 & Composition   Function 7 ($N=6$) \\
&F28 & Composition   Function 8 ($N=6$) \\
&F29 & Composition   Function 9 ($N=3$) \\
&F30 & Composition   Function 10 ($N=3$) \\
\hline
\end{tabular}
\end{table}

\subsection{Clustering-based Mutation}
For our improved mutation operator, Clu-DE first identifies a promising region in search space. This is performed, similar as in the HMS algorithm~\cite{HMS_Main_Paper}, using a clustering algorithm. We employ the well known $k$-means clustering algorithm to group the current population into $k$ clusters so that each cluster represents a region in search space. The number of clusters is selected randomly between 2 and $\sqrt{N_{P}}$~\cite{DE_Clustering01,ANN_RDE-OP}.

\begin{figure}[h!]
	\centering
	\includegraphics[width=\columnwidth]{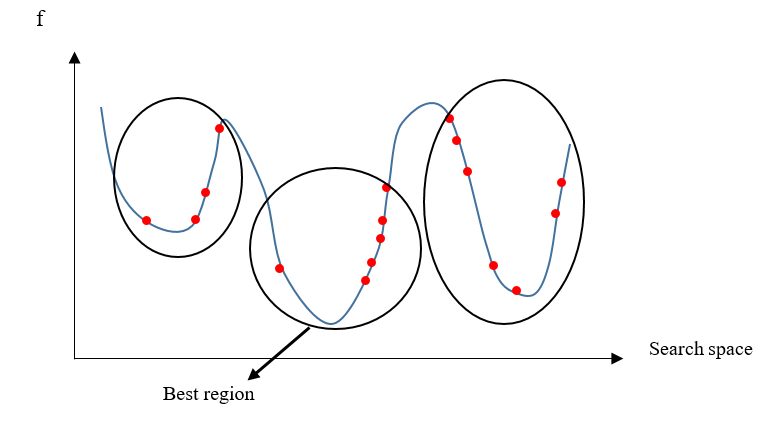}
	\caption{Population clustering in search space to identify the best region in search space (based on a minimisation problem).}
	\label{fig:searchspace}
\end{figure}

After clustering, the mean objective function value for each cluster is calculated, and the cluster with the best objective function value is then used to identify a promising region in search space. Fig.~\ref{fig:searchspace} illustrates this for a toy problem with 17 candidate solutions divided into three clusters.  

Finally, our novel clustering-based mutation is conducted as 
\begin{equation}
\overrightarrow{v^{clu}_{i}}=\overrightarrow{winner}+F(\overrightarrow{x_{r1}}-\overrightarrow{x_{r2}}) ,
\end{equation} 
where $x_{r1}$ and $x_{r2}$ are two different randomly-selected candidate solutions, and $winner$ is the best candidate solution in the promising region.  It is worth noting that the best candidate solution in the winner cluster might not be the best candidate solution in the current population. Clustering-based mutation is performed $M$ times following standard crossover and mutation.

\subsection{Population Update}
After generating $M$ new offsprings using clustering-based mutation, the population is updated for which we employ a scheme based on the generic population-based algorithm (GPBA)~\cite{Deb_population-updating}. In particular, the population is updated in the following manner:
\begin{enumerate}
	\item
	\textbf{Selection}: $k$ candidate solutions are selected randomly. This corresponds to the initial seeds for $k$-means clustering. 
	\item 
	\textbf{Generation}: $M$ new candidate solutions are created as set $v^{clu}$. This is conducted by the clustering-based mutation.
	\item
	\textbf{Replacement}: $M$ candidate solutions are selected randomly from the current population as set B.
	\item
	\textbf{Update}: From $v^{clu} \cup B$, the best $M$ individuals are selected as $\bar{B}$. The new population is then obtained as $(P-B) \cup \bar{B}$.
\end{enumerate}

\section{Experimental results}
\label{Sec:Exp}
To verify the efficacy of Clu-DE, we perform experiments on the CEC2017 benchmark functions~\cite{CEC2017}, 30 functions with different characteristics including unimodal functions, multi-modal functions, hybrid multi-modal functions, and composite functions, summarised in Table~\ref{tab:func}.

In all experiments, the maximum number of function evaluations is set to $3000 \times D$, where $D$ is the dimensionality of the search space.

The population size, crossover rate, and scaling factor are set to 50, 0.9, and 0.5, respectively. For Clu-DE, $M$ is set to 10. Each algorithm is run 25 times independently, and we report mean and standard deviation over 25 runs. 

To evaluate if there is a statistically significant difference between two algorithms, a Wilcoxon signed-rank test~\cite{tutorial_statistical} is performed with a confidence interval of 95\% on each function. 

Table~\ref{tab_D30} gives the results of Clu-DE compared to standard DE for $D=30$. From the table, we can see that Clu-DE statistically outperforms DE for 16 of the 30 functions, while obtaining equivalent performance for 12 functions. Only for two of the multi-modal functions Clu-DE yields inferior results.

\begin{table}[t!]
	\centering
	\caption{Results for $D=30$. The last column (WSRT) gives the results of the Wilcoxon signed-rank test. $+$ indicates that Clu-DE outperforms DE, $-$ the opposite, and $=$ that there is no significant difference between the two algorithms.}
	\begin{tabular}{lcccc}
		\hline
		function &  & DE & Clu-DE & WSRT \\ 
		\hline
		F1 & avg. & 8.68E+03 & 5.35E+03 & = \\
		& std.dev. & 1.47E+04 & 5.74E+03 &  \\  \hline
		F2 & avg. & 3.86E+19 & 1.97E+14 & + \\
		& std.dev. & 1.93E+20 & 9.35E+14 &  \\  \hline
		F3 & avg. & 9.43E+03 & 3.01E+02 & + \\
		& std.dev. & 8.94E+03 & 2.67E+00 &  \\  \hline
		F4 & avg. & 4.35E+02 & 4.49E+02 & = \\
		& std.dev. & 2.08E+01 & 3.14E+01 &  \\  \hline
		F5 & avg. & 6.85E+02 & 5.61E+02 & + \\
		& std.dev. & 9.19E+00 & 2.62E+01 &  \\  \hline
		F6 & avg. & 6.00E+02 & 6.00E+02 & = \\
		& std.dev. & 1.60E-04 & 2.25E-01 &  \\  \hline
		F7 & avg. & 9.12E+02 & 7.94E+02 & + \\
		& std.dev. & 1.71E+01 & 2.88E+01 &  \\  \hline
		F8 & avg. & 9.89E+02 & 8.63E+02 & + \\
		& std.dev. & 1.21E+01 & 3.03E+01 &  \\  \hline
		F9 & avg. & 9.00E+02 & 9.26E+02 & - \\
		& std.dev. & 4.94E-01 & 4.33E+01 &  \\  \hline
		F10 & avg. & 8.76E+03 & 5.89E+03 & + \\
		& std.dev. & 3.83E+02 & 9.66E+02 &  \\  \hline
		F11 & avg. & 1.13E+03 & 1.13E+03 & = \\
		& std.dev. & 2.07E+01 & 1.39E+01 &  \\  \hline
		F12 & avg. & 2.55E+05 & 7.84E+04 & + \\
		& std.dev. & 3.62E+05 & 7.16E+04 &  \\  \hline
		F13 & avg. & 1.82E+04 & 2.16E+04 & = \\
		& std.dev. & 2.15E+04 & 4.20E+04 &  \\  \hline
		F14 & avg. & 1.47E+03 & 1.44E+03 & + \\
		& std.dev. & 6.76E+00 & 1.12E+01 &  \\  \hline
		F15 & avg. & 1.61E+03 & 1.58E+03 & + \\
		& std.dev. & 8.88E+01 & 1.22E+02 &  \\  \hline
		F16 & avg. & 3.09E+03 & 2.89E+03 & + \\
		& std.dev. & 2.67E+02 & 3.33E+02 &  \\  \hline
		F17 & avg. & 2.17E+03 & 2.22E+03 & = \\
		& std.dev. & 2.01E+02 & 2.87E+02 &  \\  \hline
		F18 & avg. & 1.09E+04 & 8.24E+03 & = \\
		& std.dev. & 8.97E+03 & 7.25E+03 &  \\  \hline
		F19 & avg. & 1.92E+03 & 1.92E+03 & = \\
		& std.dev. & 5.71E+00 & 6.39E+00 &  \\  \hline
		F20 & avg. & 2.32E+03 & 2.46E+03 & - \\
		& std.dev. & 2.28E+02 & 2.20E+02 &  \\  \hline
		F21 & avg. & 2.48E+03 & 2.35E+03 & + \\
		& std.dev. & 7.34E+00 & 2.26E+01 &  \\  \hline
		F22 & avg. & 9.99E+03 & 6.52E+03 & + \\
		& std.dev. & 3.09E+02 & 2.36E+03 &  \\  \hline
		F23 & avg. & 2.83E+03 & 2.72E+03 & + \\
		& std.dev. & 1.09E+01 & 2.90E+01 &  \\  \hline
		F24 & avg. & 3.01E+03 & 2.91E+03 & + \\
		& std.dev. & 9.53E+00 & 4.32E+01 &  \\  \hline
		F25 & avg. & 2.88E+03 & 2.88E+03 & = \\
		& std.dev. & 1.16E+00 & 1.39E+00 &  \\  \hline
		F26 & avg. & 5.26E+03 & 4.30E+03 & + \\
		& std.dev. & 2.03E+02 & 2.35E+02 &  \\  \hline
		F27 & avg. & 3.20E+03 & 3.20E+03 & = \\
		& std.dev. & 1.32E-04 & 2.49E-04 &  \\  \hline
		F28 & avg. & 3.30E+03 & 3.30E+03 & = \\
		& std.dev. & 1.75E-04 & 3.47E-04 &  \\  \hline
		F29 & avg. & 3.83E+03 & 3.52E+03 & + \\
		& std.dev. & 2.48E+02 & 2.45E+02 &  \\  \hline
		F30 & avg. & 3.22E+03 & 3.22E+03 & = \\ 
		& std.dev. & 8.07E+00 & 1.90E+01 &  \\  
		\hline
		\multicolumn{4}{l}{wins/ties/losses for Clu-DE} & 16/12/2 \\ 
		\hline
	\end{tabular}
	\label{tab_D30}
\end{table}

When increasing the number of dimensions to 50, for which the results are listed in Table~\ref{tab_D50}, Clu-DE retains its efficacy. As can be seen, it statistically outperforms standard DE for 12 of the 30 functions, while giving similar results for 16 functions. 

\begin{table}[t!]
	\centering
	\caption{Results for $D=50$, laid out in same fashion as Table~\ref{tab_D30}.}
	\begin{tabular}{lcccc}
		\hline
		function &  & DE & Clu-DE & WSRT \\ 
		\hline
		F1 & avg. & 5.07E+03 & 6.25E+03 & = \\
		& std.dev. & 4.66E+03 & 9.31E+03 &  \\ \hline
		F2 & avg. & 1.02E+39 & 3.76E+42 & = \\
		& std.dev. & 4.84E+39 & 1.88E+43 &  \\ \hline
		F3 & avg. & 2.41E+05 & 8.26E+03 & + \\
		& std.dev. & 5.65E+04 & 4.56E+03 &  \\ \hline
		F4 & avg. & 4.49E+02 & 4.87E+02 & - \\
		& std.dev. & 2.42E+01 & 4.00E+01 &  \\ \hline
		F5 & avg. & 8.61E+02 & 6.23E+0+ & + \\
		& std.dev. & 2.04E+01 & 3.77E+01 &  \\ \hline
		F6 & avg. & 6.00E+02 & 6.00E+02 & = \\
		& std.dev. & 7.71E-02 & 1.41E+00 &  \\ \hline
		F7 & avg. & 1.12E+03 & 9.40E+02 & + \\
		& std.dev. & 1.63E+01 & 5.26E+01 &  \\ \hline
		F8 & avg. & 1.15E+03 & 9.20E+02 & + \\
		& std.dev. & 6.57E+01 & 3.22E+01 &  \\ \hline
		F9 & avg. & 9.09E+02 & 1.44E+03 & - \\
		& std.dev. & 1.04E+01 & 4.37E+02 &  \\ \hline
		F10 & avg. & 1.54E+04 & 9.73E+03 & + \\
		& std.dev. & 3.99E+02 & 1.57E+03 &  \\ \hline
		F11 & avg. & 1.20E+03 & 1.21E+03 & = \\
		& std.dev. & 6.55E+01 & 3.32E+01 &  \\ \hline
		F12 & avg. & 1.96E+06 & 1.89E+06 & = \\
		& std.dev. & 1.47E+06 & 1.17E+06 &  \\ \hline
		F13 & avg. & 1.13E+04 & 2.88E+04 & = \\
		& std.dev. & 1.39E+04 & 4.45E+04 &  \\ \hline
		F14 & avg. & 7.39E+03 & 1.14E+04 & = \\
		& std.dev. & 1.20E+04 & 1.55E+04 &  \\ \hline
		F15 & avg. & 3.24E+04 & 1.73E+04 & = \\
		& std.dev. & 4.17E+04 & 2.24E+04 &  \\ \hline
		F16 & avg. & 4.69E+03 & 4.08E+03 & + \\
		& std.dev. & 5.37E+02 & 7.70E+02 &  \\ \hline
		F17 & avg. & 3.34E+03 & 3.30E+03 & = \\
		& std.dev. & 3.78E+02 & 3.67E+02 &  \\ \hline
		F18 & avg. & 1.09E+05 & 5.80E+04 & + \\
		& std.dev. & 6.76E+04 & 3.75E+04 &  \\ \hline
		F19 & avg. & 1.00E+04 & 1.07E+04 & = \\
		& std.dev. & 1.39E+04 & 8.79E+03 &  \\ \hline
		F20 & avg. & 3.47E+03 & 3.55E+03 & = \\
		& std.dev. & 3.53E+02 & 4.40E+02 &  \\ \hline
		F21 & avg. & 2.66E+03 & 2.41E+03 & + \\
		& std.dev. & 1.70E+01 & 3.07E+01 &  \\ \hline
		F22 & avg. & 1.67E+04 & 1.24E+04 & + \\
		& std.dev. & 3.92E+02 & 1.79E+03 &  \\ \hline
		F23 & avg. & 3.07E+03 & 2.90E+03 & + \\
		& std.dev. & 3.71E+01 & 6.85E+01 &  \\ \hline
		F24 & avg. & 3.28E+03 & 3.06E+03 & + \\
		& std.dev. & 1.60E+01 & 5.22E+01 &  \\ \hline
		F25 & avg. & 2.94E+03 & 2.97E+03 & = \\
		& std.dev. & 2.32E+01 & 3.21E+01 &  \\ \hline
		F26 & avg. & 7.35E+03 & 5.35E+03 & = \\
		& std.dev. & 4.57E+02 & 5.20E+02 &  \\ \hline
		F27 & avg. & 3.20E+03 & 3.20E+03 & = \\
		& std.dev. & 1.30E-04 & 4.01E-04 &  \\ \hline
		F28 & avg. & 3.30E+03 & 3.30E+03 & = \\
		& std.dev. & 1.67E-04 & 4.20E-04 &  \\ \hline
		F29 & avg. & 5.04E+03 & 4.23E+03 & + \\
		& std.dev. & 3.66E+02 & 4.76E+02 &  \\ \hline
		F30 & avg. & 7.09E+03 & 7.48E+03 & = \\
		& std.dev. & 4.81E+03 & 4.70E+03 &  \\ 
		\hline
		\multicolumn{4}{l}{wins/ties/losses for Clu-DE} & 12/16/2 \\ 
		\hline
	\end{tabular}
	\label{tab_D50}
\end{table}

For $D=100$, the results are given in Table~\ref{tab_D100}. As we can see from there, Clu-DE obtains better or similar results for 24 of the 30 functions, thus clearly outperforming DE also for high-dimensional problems.

\begin{table}[t!]
	\centering
	\caption{Results for $D=100$, laid out in same fashion as Table~\ref{tab_D30}.}
	\begin{tabular}{lcccc}
		\hline
		function &  & DE & Clu-DE & WSRT \\ \hline
		F1 & avg. & 9.77E+03 & 4.00E+03 & + \\
		& std.dev. & 1.21E+04 & 6.94E+03 &  \\ \hline
		F2 & avg. & 7.36E+92 & 2.29E+105 & - \\
		& std.dev. & 3.68E+93 & 1.14E+106 &  \\ \hline
		F3 & avg. & 1.58E+06 & 1.54E+05 & + \\
		& std.dev. & 3.59E+05 & 3.85E+04 &  \\ \hline
		F4 & avg. & 5.94E+02 & 6.57E+02 & = \\
		& std.dev. & 5.55E+01 & 4.92E+01 &  \\ \hline
		F5 & avg. & 1.21E+03 & 8.95E+02 & + \\
		& std.dev. & 3.11E+02 & 7.70E+01 &  \\ \hline
		F6 & avg. & 6.01E+02 & 6.13E+02 & - \\
		& std.dev. & 4.44E-01 & 3.06E+00 &  \\ \hline
		F7 & avg. & 1.74E+03 & 1.48E+03 & + \\
		& std.dev. & 4.17E+01 & 1.52E+02 &  \\ \hline
		F8 & avg. & 1.51E+03 & 1.18E+03 & + \\
		& std.dev. & 3.02E+02 & 1.20E+02 &  \\ \hline
		F9 & avg. & 1.91E+03 & 1.00E+04 & - \\
		& std.dev. & 1.49E+03 & 4.80E+03 &  \\ \hline
		F10 & avg. & 3.28E+04 & 2.29E+04 & + \\
		& std.dev. & 5.47E+02 & 2.92E+03 &  \\ \hline
		F11 & avg. & 3.50E+03 & 1.61E+03 & + \\
		& std.dev. & 1.22E+03 & 2.39E+02 &  \\ \hline
		F12 & avg. & 6.72E+06 & 1.20E+07 & - \\
		& std.dev. & 4.00E+06 & 6.59E+06 &  \\ \hline
		F13 & avg. & 7.94E+03 & 9.60E+03 & = \\
		& std.dev. & 9.64E+03 & 1.31E+04 &  \\ \hline
		F14 & avg. & 4.64E+05 & 4.33E+05 & = \\
		& std.dev. & 3.51E+05 & 2.82E+05 &  \\ \hline
		F15 & avg. & 6.26E+03 & 8.32E+03 & = \\
		& std.dev. & 6.46E+03 & 9.75E+03 &  \\ \hline
		F16 & avg. & 1.01E+04 & 7.18E+03 & + \\
		& std.dev. & 3.67E+02 & 1.56E+03 &  \\ \hline
		F17 & avg. & 7.00E+03 & 6.15E+03 & + \\
		& std.dev. & 7.39E+02 & 8.37E+02 &  \\ \hline
		F18 & avg. & 9.74E+05 & 5.34E+05 & + \\
		& std.dev. & 4.26E+05 & 3.62E+05 &  \\ \hline
		F19 & avg. & 3.88E+03 & 5.25E+03 & = \\
		& std.dev. & 3.09E+03 & 4.40E+03 &  \\ \hline
		F20 & avg. & 7.07E+03 & 6.61E+03 & + \\
		& std.dev. & 7.01E+02 & 7.97E+02 &  \\ \hline
		F21 & avg. & 3.08E+03 & 2.72E+03 & + \\
		& std.dev. & 2.74E+02 & 6.93E+01 &  \\ \hline
		F22 & avg. & 3.46E+04 & 2.57E+04 & + \\
		& std.dev. & 5.29E+02 & 2.41E+03 &  \\ \hline
		F23 & avg. & 3.05E+03 & 3.32E+03 & - \\
		& std.dev. & 3.71E+01 & 6.19E+01 &  \\ \hline
		F24 & avg. & 4.07E+03 & 3.97E+0= & = \\
		& std.dev. & 2.76E+02 & 1.05E+02 &  \\ \hline
		F25 & avg. & 3.26E+03 & 3.29E+03 & = \\
		& std.dev. & 7.68E+01 & 5.90E+01 &  \\ \hline
		F26 & avg. & 1.13E+04 & 1.37E+04 & - \\
		& std.dev. & 3.48E+03 & 1.12E+03 &  \\ \hline
		F27 & avg. & 3.20E+03 & 3.20E+03 & = \\
		& std.dev. & 1.42E-04 & 3.31E-04 &  \\ \hline
		F28 & avg. & 3.30E+03 & 3.30E+03 & = \\
		& std.dev. & 7.72E-05 & 1.11E+01 &  \\ \hline
		F29 & avg. & 8.30E+03 & 6.81E+03 & + \\
		& std.dev. & 8.11E+02 & 9.69E+02 &  \\ \hline
		F30 & avg. & 1.02E+04 & 1.50E+04 & = \\
		& std.dev. & 9.45E+03 & 1.86E+04 &  \\ 
		\hline
		\multicolumn{4}{l}{wins/ties/losses for Clu-DE} & 14/10/6 \\ 
		\hline
	\end{tabular}
	\label{tab_D100}
\end{table}

\begin{figure}[t!]
	\centering
	\begin{subfigure}[b]{0.23\textwidth}
		\centering
		\includegraphics[width=\textwidth]{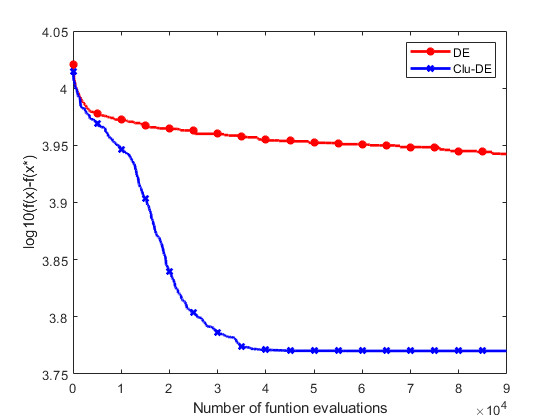}
		\caption{F10, $D=30$}
	\end{subfigure}
	\hfill
	\begin{subfigure}[b]{0.23\textwidth}
		\centering
		\includegraphics[width=\textwidth]{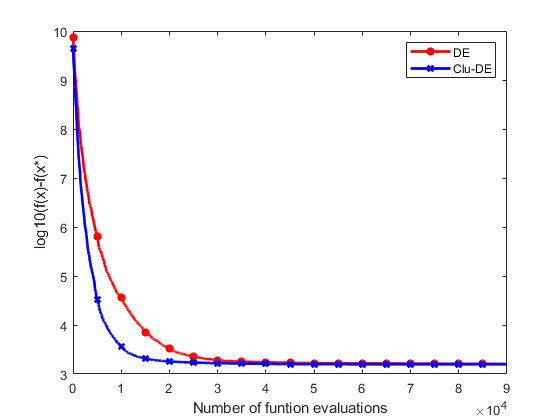}
		\caption{F15, $D=30$}
	\end{subfigure}
	
	\begin{subfigure}[b]{0.23\textwidth}
		\centering
		\includegraphics[width=\textwidth]{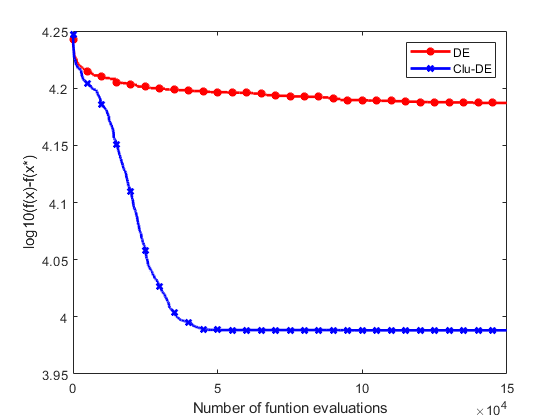}
		\caption{F10, $D=50$}
	\end{subfigure}	
	\hfill
	\begin{subfigure}[b]{0.23\textwidth}
		\centering
		\includegraphics[width=\textwidth]{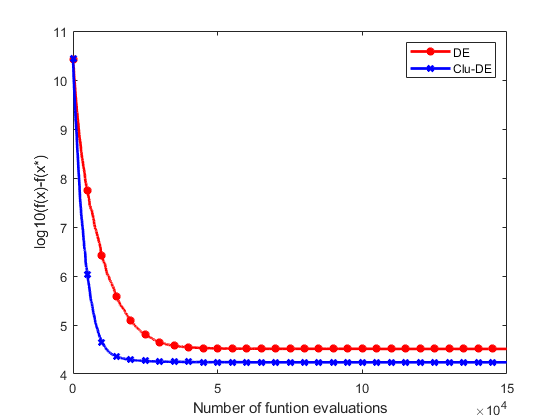}
		\caption{F15, $D=50$}
	\end{subfigure}
	
	\begin{subfigure}[b]{0.23\textwidth}
		\centering
		\includegraphics[width=\textwidth]{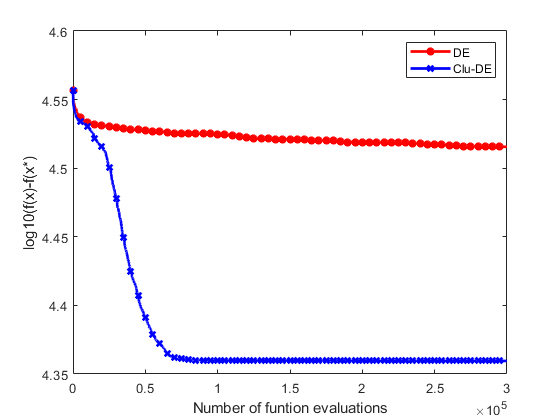}
		\caption{F10, $D=100$}
	\end{subfigure}
	\hfill
	\begin{subfigure}[b]{0.23\textwidth}
		\centering
		\includegraphics[width=\textwidth]{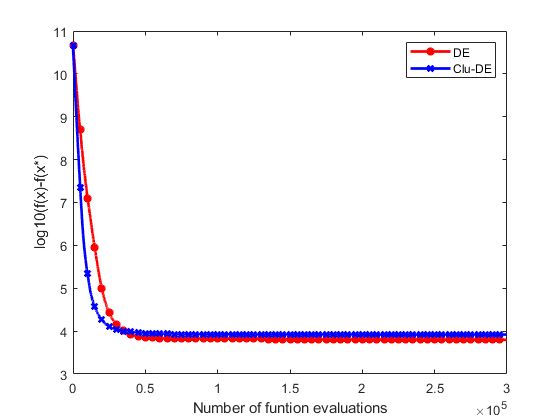}
		\caption{F15,D=100}
	\end{subfigure}
	
	\caption{Convergence plots for F10 and F15.} 
	\label{fig:Convergence_Curve_2017}
\end{figure}

Last but not least, Figure~\ref{fig:Convergence_Curve_2017} shows convergence curves of our proposed algorithm compared to DE for, as representative examples, F10 and F15 and all dimensionalities. As we can observe, Clu-DE converges faster than standard DE.

\section{Conclusions}
\label{Sec:conc}
In this paper, we have proposed a novel differential evolution algorithm, Clu-DE, based on a novel clustering-based mutation operator. A promising region in search space is found using $k$-means clustering and some new candidate solutions are generated using the proposed cluster-based mutation. A population update scheme is introduced to include the new candidate solutions into the current population. Extensive experiments on the CEC-2017 benchmark functions and for dimensionalities of 30, 50 and 100 verify that Clu-DE is a competitive variant of DE. In future work, we intend to extend Clu-DE for multi-objective optimisation problems.

\IEEEtriggeratref{18} 

\bibliographystyle{IEEEtran}
\bibliography{jalal}

\begin{thebibliography}{10}
\providecommand{\url}[1]{#1}
\csname url@samestyle\endcsname
\providecommand{\newblock}{\relax}
\providecommand{\bibinfo}[2]{#2}
\providecommand{\BIBentrySTDinterwordspacing}{\spaceskip=0pt\relax}
\providecommand{\BIBentryALTinterwordstretchfactor}{4}
\providecommand{\BIBentryALTinterwordspacing}{\spaceskip=\fontdimen2\font plus
\BIBentryALTinterwordstretchfactor\fontdimen3\font minus
  \fontdimen4\font\relax}
\providecommand{\BIBforeignlanguage}[2]{{%
\expandafter\ifx\csname l@#1\endcsname\relax
\typeout{** WARNING: IEEEtran.bst: No hyphenation pattern has been}%
\typeout{** loaded for the language `#1'. Using the pattern for}%
\typeout{** the default language instead.}%
\else
\language=\csname l@#1\endcsname
\fi
#2}}
\providecommand{\BIBdecl}{\relax}
\BIBdecl

\bibitem{GHMS-RCS}
S.~J. Mousavirad, G.~Schaefer, and I.~Korovin, ``A global-best guided human
  mental search algorithm with random clustering strategy,'' in
  \emph{International Conference on Systems, Man and Cybernetics}, 2019, pp.
  3174--3179.

\bibitem{PSO_Main_Paper02}
J.~Kennedy and R.~Eberhart, ``Particle swarm optimization ({PSO}),'' in
  \emph{IEEE International Conference on Neural Networks}, 1995, pp.
  1942--1948.

\bibitem{center_sampling_PSO_SMC2020}
S.~J. Mousavirad and S.~Rahnamayan, ``{CenPSO}: A novel center-based particle
  swarm optimization algorithm for large-scale optimization,'' in
  \emph{International Conference on Systems, Man, and Cybernetics}, 2020.

\bibitem{DE_Original}
R.~Storn and K.~Price, ``Differential evolution--a simple and efficient
  heuristic for global optimization over continuous spaces,'' \emph{Journal of
  Global Optimization}, vol.~11, no.~4, pp. 341--359, 1997.

\bibitem{Image_Segmentation_auto_DE}
S.~Das and A.~Konar, ``Automatic image pixel clustering with an improved
  differential evolution,'' \emph{Applied Soft Computing}, vol.~9, no.~1, pp.
  226--236, 2009.

\bibitem{Image_Thresholding_CenDE}
S.~J. Mousavirad, S.~Rahnamayan, and G.~Schaefer, ``Many-level image
  thresholding using a center-based differential evolution algorithm,'' in
  \emph{Congress on Evolutionary Computation}, 2020.

\bibitem{ANN_DE_opposition}
S.~J. Mousavirad and S.~Rahnamayan, ``Evolving feedforward neural networks
  using a quasi-opposition-based differential evolution for data
  classification,'' in \emph{IEEE Symposium Series on Computational
  Intelligence}, 2020.

\bibitem{ANN_RDE-OP}
S.~J. Mousavirad, G.~Schaefer, I.~Korovin, and D.~Oliva, ``{RDE-OP}: A
  region-based differential evolution algorithm incorporation opposition-based
  learning for optimising the learning process of multi-layer neural
  networks,'' in \emph{24th International Conference on the Applications of
  Evolutionary Computation}, 2021.

\bibitem{ANN_DE_cen_Op}
N.~H. Awad, M.~Z. Ali, P.~N. Suganthan, and R.~G. Reynolds, ``Differential
  evolution-based neural network training incorporating a centroid-based
  strategy and dynamic opposition-based learning,'' in \emph{IEEE Congress on
  Evolutionary Computation}, 2016, pp. 2958--2965.

\bibitem{DE_financial01}
A.~Ara, N.~A. Khan, O.~A. Razzaq, T.~Hameed, and M.~A.~Z. Raja, ``Wavelets
  optimization method for evaluation of fractional partial differential
  equations: an application to financial modelling,'' \emph{Advances in
  Difference Equations}, vol. 2018, no.~1, p.~8, 2018.

\bibitem{DE_financial02}
Y.~Tang, J.~Ji, Y.~Zhu, S.~Gao, Z.~Tang, and Y.~Todo, ``A differential
  evolution-oriented pruning neural network model for bankruptcy prediction,''
  \emph{Complexity}, vol. 2019, 2019.

\bibitem{CI-DE}
J.~Feng, J.~Zhang, C.~Wang, and M.~Xu, ``Self-adaptive collective
  intelligence-based mutation operator for differential evolution algorithms,''
  \emph{The Journal of Supercomputing}, vol.~76, no.~2, pp. 876--896, 2020.

\bibitem{DE_ensemble02}
G.~Wu, R.~Mallipeddi, P.~N. Suganthan, R.~Wang, and H.~Chen, ``Differential
  evolution with multi-population based ensemble of mutation strategies,''
  \emph{Information Sciences}, vol. 329, pp. 329--345, 2016.

\bibitem{CODE_01}
Y.~Wang, Z.~Cai, and Q.~Zhang, ``Differential evolution with composite trial
  vector generation strategies and control parameters,'' \emph{IEEE
  Transactions on Evolutionary Computation}, vol.~15, no.~1, pp. 55--66, 2011.

\bibitem{DE_tournament}
D.~Bajer, ``Adaptive k-tournament mutation scheme for differential evolution,''
  \emph{Applied Soft Computing}, vol.~85, p. 105776, 2019.

\bibitem{DE_neighborhood}
B.-Y. Qu, P.~N. Suganthan, and J.-J. Liang, ``Differential evolution with
  neighborhood mutation for multimodal optimization,'' \emph{IEEE Transactions
  on Evolutionary Computation}, vol.~16, no.~5, pp. 601--614, 2012.

\bibitem{DE_Competition_ICCSE2019}
S.~J. Mousavirad and S.~Rahnamayan, ``Differential evolution algorithm based on
  a competition scheme,'' in \emph{14th International Conference on Computer
  Science and Education}, 2019.

\bibitem{HMS_Main_Paper}
S.~J. Mousavirad and H.~Ebrahimpour-Komleh, ``Human mental search: a new
  population-based metaheuristic optimization algorithm,'' \emph{Applied
  Intelligence}, vol.~47, no.~3, pp. 850--887, 2017.

\bibitem{k-means_Original}
J.~MacQueen, ``Some methods for classification and analysis of multivariate
  observations,'' in \emph{5th Berkeley Symposium on Mathematical Statistics
  and Probability}, 1967, pp. 281--297.

\bibitem{CEC2017}
G.~Wu, R.~Mallipeddi, and P.~Suganthan, ``Problem definitions and evaluation
  criteria for the {CEC 2017} competition on constrained real-parameter
  optimization,'' Nanyang Technological University, Singapore, Tech. Rep.,
  2016.

\bibitem{DE_Clustering01}
Z.~Cai, W.~Gong, C.~X. Ling, and H.~Zhang, ``A clustering-based differential
  evolution for global optimization,'' \emph{Applied Soft Computing}, vol.~11,
  no.~1, pp. 1363--1379, 2011.

\bibitem{Deb_population-updating}
K.~Deb, ``A population-based algorithm-generator for real-parameter
  optimization,'' \emph{Soft Computing}, vol.~9, no.~4, pp. 236--253, 2005.

\bibitem{tutorial_statistical}
J.~Derrac, S.~Garc{\'\i}a, D.~Molina, and F.~Herrera, ``A practical tutorial on
  the use of nonparametric statistical tests as a methodology for comparing
  evolutionary and swarm intelligence algorithms,'' \emph{Swarm and
  Evolutionary Computation}, vol.~1, no.~1, pp. 3--18, 2011.

\end{thebibliography}

\end{document}